\pdfoutput=1

\documentclass[11pt]{article}

\usepackage[]{acl2023}

\usepackage{times}
\usepackage{latexsym}

\usepackage[T1]{fontenc}

\usepackage[utf8]{inputenc}

\usepackage{microtype}

\usepackage{amssymb}
\usepackage{pifont}%
\newcommand{\cmark}{\ding{51}}%
\newcommand{\xmark}{\ding{55}}%

\usepackage{amsmath}
\usepackage{booktabs}
\usepackage{subfig}
\usepackage{tikz}
\usetikzlibrary{arrows,fit,positioning,calc}
\usepackage{pgfplots}
\pgfplotsset{width=7.5cm,compat=1.16}
\usepackage{siunitx}
\makeatletter
\pgfdeclareshape{document}{
\inheritsavedanchors[from=rectangle] %
\inheritanchorborder[from=rectangle]
\inheritanchor[from=rectangle]{center}
\inheritanchor[from=rectangle]{north}
\inheritanchor[from=rectangle]{south}
\inheritanchor[from=rectangle]{west}
\inheritanchor[from=rectangle]{east}
\backgroundpath{%
\southwest \pgf@xa=\pgf@x \pgf@ya=\pgf@y
\northeast \pgf@xb=\pgf@x \pgf@yb=\pgf@y
\pgf@xc=\pgf@xb \advance\pgf@xc by-10pt %
\pgf@yc=\pgf@yb \advance\pgf@yc by-10pt
\pgfpathmoveto{\pgfpoint{\pgf@xa}{\pgf@ya}}
\pgfpathlineto{\pgfpoint{\pgf@xa}{\pgf@yb}}
\pgfpathlineto{\pgfpoint{\pgf@xc}{\pgf@yb}}
\pgfpathlineto{\pgfpoint{\pgf@xb}{\pgf@yc}}
\pgfpathlineto{\pgfpoint{\pgf@xb}{\pgf@ya}}
\pgfpathclose
\pgfpathmoveto{\pgfpoint{\pgf@xc}{\pgf@yb}}
\pgfpathlineto{\pgfpoint{\pgf@xc}{\pgf@yc}}
\pgfpathlineto{\pgfpoint{\pgf@xb}{\pgf@yc}}
\pgfpathlineto{\pgfpoint{\pgf@xc}{\pgf@yc}}
}
}
\makeatother

\title{RISE: Leveraging Retrieval Techniques for Summarization Evaluation}

\author{David Uthus \\ Google Research \\ \tt \small duthus@google.com
\And
Jianmo Ni \\ Google Deepmind \\ \tt \small jianmon@google.com}

\begin{document}
\maketitle
\begin{abstract}
Evaluating automatically-generated text summaries is a challenging task.
While there have been many interesting approaches, they still fall short of human evaluations.
We present RISE, a new approach for evaluating summaries by leveraging techniques from information retrieval.
RISE is first trained as a retrieval task using a dual-encoder retrieval setup, and can then be subsequently utilized for evaluating a generated summary given an input document, without gold reference summaries.
RISE is especially well suited when working on new datasets where one may not have reference summaries available for evaluation.
We conduct comprehensive experiments on the SummEval benchmark \citep{fabbri-2021} and a long document summarization benchmark. The results show that RISE consistently achieves higher correlation with human evaluations compared to many past approaches to summarization evaluation. 
Furthermore, RISE also demonstrates data-efficiency and generalizability across languages. 
\end{abstract}

\section{Introduction}

Summarization evaluation has became a topic of interest in recent years.
In the past, many summarization approaches have relied on ROUGE \cite{lin-2004-rouge} for evaluating generated summaries.
Yet, as reported by \citet{fabbri-2021}, ROUGE and other automated metrics tend to fall short when compared to human evaluations.
To overcome this, many new approaches have been developed leveraging pre-trained language models, showing various degrees of success.

We present our new approach to summarization evaluation called Retrieval-Inspired Summarization Evaluation (RISE).
As with recent approaches, RISE leverages pre-trained language models.
But unlike past approaches, we treat evaluation as a retrieval task, leveraging techniques from information retrieval.
This is done by using a dual-encoder approach \citep{gillick-etal-2019-learning,ni2021large}, feeding in the source document and the summary to be evaluated, in order to get a final score for evaluation.

The benefits of RISE are as follows:

\begin{itemize}
\item Our experiments show that RISE strongly correlates with human metrics, outperforming many recent approaches.
It works well when fine-tuned with in-domain data, when transferred to new domains, transferred to new languages, or on a small amount of data.

\item It has the benefit of not being reliant on reference summaries during evaluation or output calibration. This allows it to work well in new domains or online use cases, where it may be expensive or impractical to obtain reference summaries.

\item RISE can be further improved as better pretrained language models are released in the future.
We have released checkpoints and code to evaluate with our models and train users' own evaluation models.\footnote{\url{https://github.com/google-research/google-research/tree/master/rise}}
\end{itemize} 

\begin{table}
\small
\centering
\begin{tabular}{l|c|c|c}
\toprule
Method & Source- & Reference- & Model-  \\
       & free & free & based  \\
\midrule
ROUGE & \cmark  & \xmark  & \xmark   \\
\textsc{chrF}   & \cmark  & \xmark  & \xmark   \\
BERTScore & \cmark & \xmark & \cmark   \\
SMART & \xmark & \xmark  &  \cmark  \\
T5-ANLI & \xmark & \cmark & \cmark  \\
BARTScore & \xmark  & \cmark  & \cmark  \\
RISE & \xmark & \cmark  &  \cmark  \\
\bottomrule
\end{tabular}
\caption{
Comparisons of different summarization evaluation methods. Source-free methods often requires golden summaries as reference for evaluation; meanwhile reference-free methods only rely on the source input, which is more practical when gold summaries are hard to obtain.}
\label{tab:eval_methods}
\end{table}

\begin{figure*}[htp]
\centering

\tikzstyle{doc}=[%
draw,
thick,
align=center,
color=black,
fill=orange!20,
shape=document,
minimum width=20mm,
minimum height=28.2mm,
shape=document,
inner sep=2ex,
]

\subfloat{%
\begin{tikzpicture}
   [
  font=\sffamily,
  every matrix/.style={ampersand replacement=\&,column sep=.5cm,row sep=.5cm},
  encoder/.style={draw,thick,rounded corners,fill=yellow!20,inner sep=.3cm},
  process/.style={draw,thick,circle,fill=cyan!20},
  summary/.style={draw,thick,fill=green!20,inner sep=.3cm},
  outline/.style={draw,thick,fill=gray!20,inner sep=.3cm},
  to/.style={->,>=stealth',shorten >=1pt,semithick,font=\sffamily\footnotesize},
  every node/.style={align=center}]
  \small
  \node[doc] (doc) {Source \\ \scriptsize Fleetwood are the only team still \\ \scriptsize to have a 100\% record in \\ \scriptsize Sky Bet League One as a 2-0 win \\ \scriptsize over Scunthorpe sent Graham \\ \scriptsize Alexander’s  men top of the table...};
  
  \node[summary,right=of doc.east, yshift=0cm] (summary) {Summary \\ \scriptsize Fleetwood top of League One \\ \scriptsize after 2-0 win at Scunthorpe.};
  \node[below=of summary.south, yshift=-.2cm, xshift=.4cm] (outline3) {In-batch Negatives};
  \node[outline,below=of outline3.west, yshift=.1cm, anchor=west] (o3) {\scriptsize Prime Minister and \\ \scriptsize his family...};
  \node[left=of outline3.west, xshift=.5cm] (outline2) {Model Negatives};
  \node[outline,below=of outline2.west, yshift=.1cm, anchor=west] (o2) {\scriptsize Bristol City remain \\ \scriptsize a point behind...};
  \node[left=of outline2.west, xshift=.5cm] (outline1) {Lexical Negatives};
  \node[outline,below=of outline1.west, yshift=.1cm, anchor=west] (o1) {\scriptsize Scunthorpe top of \\ \scriptsize League One...};

  \node [draw=black!50, fit={(outline1) (o1)}] {};
  \node [draw=black!50, fit={(outline2) (o2)}] {};
  \node [draw=black!50, fit={(outline3) (o3)}] {};
  
  \draw[to] (doc) -- (summary);

\end{tikzpicture}
}%
\hfill
\subfloat{%
\begin{tikzpicture}
  [
  font=\sffamily,
  every matrix/.style={ampersand replacement=\&,column sep=-.5cm,row sep=.5cm},
  encoder/.style={draw,thick,rounded corners,fill=yellow!20,inner sep=.3cm},
  process/.style={draw,thick,circle,fill=cyan!20},
  source/.style={draw,thick,fill=orange!20,inner sep=.3cm},
  summary/.style={draw,thick,fill=green!20,inner sep=.3cm},
  to/.style={->,>=stealth',shorten >=1pt,semithick,font=\sffamily\footnotesize},
  every node/.style={align=center}]
  \small

  \matrix{
    \& \node[process] (s) {Eval score \\ $(h_x)^T \cdot h_y$};   \& \\
    \node[encoder] (e1) {T5 encoder\\$h_x$}; \&  \& \node[encoder] (e2) {T5 encoder\\$h_y$}; \\
    \node[source] (input) {(input) \\ source}; \&  \& \node[summary] (target) {(target) \\ summary}; \\
  };

  \draw[to] (input) -- (e1);
  \draw[to] (target) -- (e2);
  \draw[to] (e1) -- (s);
  \draw[to] (e2) -- (s);

\end{tikzpicture}
}%
\caption{Diagram of RISE.
RISE first depends on an source document and a summary.
As part of the training process, it will also require example negatives.
These can be either in-batch negatives, lexical negatives (via augmenting the original summary), or model negatives (via mining for similar negatives with a trained model).
When evaluating a summary, RISE takes in the source document as input and summary to be evaluated as the target.
It will encode both input and summary with a T5 encoder. Finally, it will take the dot product of these encodings, resulting in the score we use for evaluating the summary.
}
\label{fig:rise}
\end{figure*}
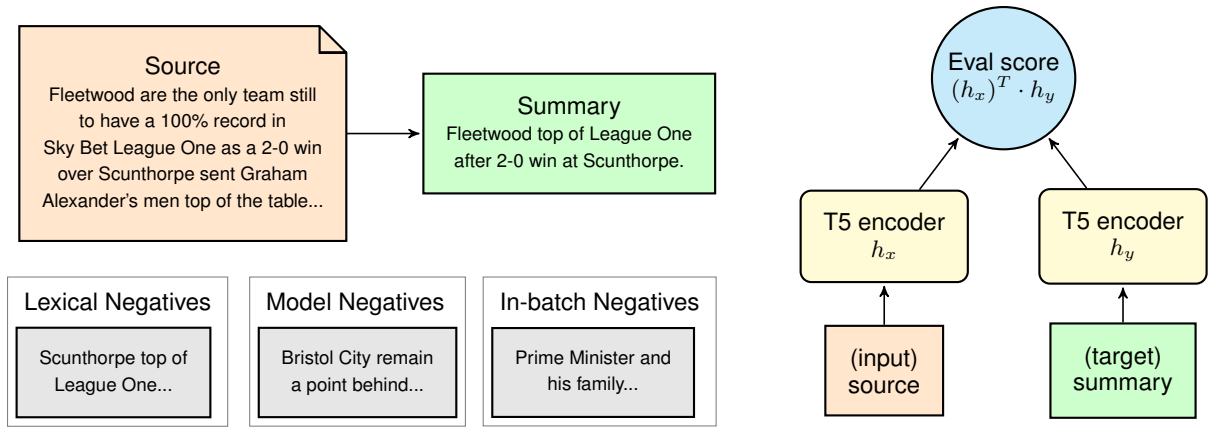

\section{Related Work}

There has been a lot of work in recent years on evaluation of summarization.
These approaches have been diverse, and there are different ways of categorizing them.
For example, \citet{yuan-2021-bartscore} grouped evaluation metrics into four categories: matching, regressions, ranking and generation.
In this case, RISE falls under ranking.
One of the idealistic benefits of creating such a ranking approach is that while we need references during training of such a model, for evaluation we can focus on just the source document and the generated summary to be evaluated.

In this paper, we are more focused on metrics that are reference-free or reference-dependent.
Table \ref{tab:eval_methods} shows how we can group such metrics.
Note that some of the metrics were designed specifically for summarization, while others were designed for generation in general.
This is also not an exhaustive list of all metrics.

Some of the metrics rely on comparing a generated summary with a reference summary, such as ROUGE \cite{lin-2004-rouge}, \textsc{chrF} \cite{popovic-2015-chrf}, and BERTScore \cite{zhang-2020-bertscore}.
Others rely on comparing the generated summary with the source document, such as T5-ANLI \cite{honovich-etal-2022-true-evaluating}, BARTScore \cite{yuan-2021-bartscore}, and the work we present here, RISE.
SMART \cite{amplayo-2022-smart}, unlike these other metrics, can compare with both the source and reference when computing metrics.

More recently, BARTScore has been proposed as a competitive reference-free evaluation method that leverages the power of pre-trained language models. Given an input document, it computes the likelihood of generating the summary from the BART model and then use the likelihood as the quality score. 
The benefits of this approach is that it can work with a pre-trained language model without requiring any finetuning, though finetuning does help improve its performance for specific datasets.
A drawback of this approach is that the metric scores are challenging to interpret, making them less practical to use for quality filtering or calibration of summarization models.

The benefit of being a reference-free approach is that it can work well when one needs to evaluate a new generated summary where there is no reference summary available.
But doing so is a more challenging task, especially when needing to handle longer inputs of source documents (which can be further challenging when the source document is very long).

\section{Model}

Figure \ref{fig:rise} shows how RISE works.
As previously mentioned, RISE leverages pre-trained language models via a dual-encoder network.
To do so, RISE builds upon T5X Retrieval\footnote{\footnotesize \url{https://github.com/google-research/t5x_retrieval}} as the framework for the dual-encoder model.
Each encoder in the model is the encoder of a pre-trained T5 model \cite{c4}.
T5 is an encoder-decoder model, thus we are only using the encoder half of the model.

To evaluate a summary, RISE will feed in the source document $d_i$ into one of the encoders (the left encoder in Figure \ref{fig:rise}), and the summary $s_i$ to be evaluated into the other encoder.
Both documents are encoded, and finally we take the dot product of the resultant output encodings in order to get the final score used for evaluation.

To train the dual-encoder models, we apply contrastive learning and use the positive pairs of (document, summary). Assume $s_i^{+}$ is considered as a positive summary for document $d_i$. During training, all other summaries in a batch are considered as negatives.
The models are trained using an in-batch sampled softmax~\citep{Henderson2017EfficientNL}:
\begin{equation}
    \mathcal{L} = -\mathrm{log}\frac{e^{\text{sim}(d_i, s_i^{+})/ \tau}}{\sum_{j \in \mathcal{B}} { e^{\text{sim}(d_i, s_j^{+}) / \tau} } }.
    \label{eq::loss}
\end{equation}
The similarity scoring function \textit{sim} is the cosine distance in our experiments\footnote{Specifically, we apply l2-normalization to the document and summary encodings, then compute their dot-product.}. At inference time, the similarity scores are used to estimate the quality of generated summaries.
$\mathcal{B}$ is a mini-batch of examples and $\tau$ is the softmax temperature. During training, we could prepare additional negatives $s_i^{-}$ for each document $d_i$, and the loss can be computed as:
\begin{equation}
    \mathcal{L} = -\mathrm{log}\frac{e^{\text{sim}(d_i, s_i^{+})/ \tau }}{\sum_{j \in \mathcal{B}} { e^{\text{sim}(d_i, s_j^{+})/ \tau}  + e^{\text{sim}(d_i, s_i^{-})/ \tau} }} .
\end{equation}

By building on top of T5 architecture with T5X Retrieval \citep{ni2021large}, this gives us the benefit of being able to leverage different pre-trained T5 models.
This includes the original T5 model for tasks with shorter inputs, LongT5 \cite{guo-etal-2022-longt5} for tasks with longer inputs, and mT5 \cite{xue-etal-2021-mt5} for multilingual tasks.

\subsection{Training}

RISE, being built upon T5X Retrieval, needs to be finetuned so that it can learn to score a summary given an input source document.
To do so, we can make use of various summarization datasets for finetuning.
There are three strategies for finetuning that we have explored:

\begin{itemize}
  \item Training only on in-domain data - can see how well the model performs if only trained on the targeted task.

  \item Training only on out-of-domain data - can see how well the model transfers to new domains.

  \item Training on both in- and out-domain data - see how well the model performs if trained on a mixture of data outside the domain and within the domain.
\end{itemize}

\subsection{Generating Negatives}

Given a suitable dataset, we then can finetune RISE as a retrieval task.
As RISE is being trained as a retrieval model, it requires example negatives during training to help differentiate the correct summary to ``retrieve'' given a set of summaries.
By default, when finetuning one can use other summaries within a batch as candidate summaries.
Additionally, a model can be trained with additional hard negatives per example.
We looked at three possibilities for generating these (illustrated in Figure \ref{fig:rise}): lexical negatives, model negatives, or a combination of both.

\subsubsection{Lexical Negatives}

Lexical negatives are example negatives generated by augmenting a reference summary.
The goal is to augment the summary in such a way that it exhibits characteristics that make it a poor summary compared to the reference.
This would then allow the model to learn to differentiate a good summary from a poor summary.

For data augmentations, we looked at several methods for augmenting the original summary:

\begin{itemize}
\item Swapping noun entities - randomly swap noun entities in the summary with one from the original source document.

\item Shuffling words - randomly shuffle the words in the summary.

\item Dropping words - randomly drop words from the summary.

\item Dropping characters - randomly drop characters from the summary.

\item Swapping antonyms - swap words with their antonyms.
\end{itemize}

One benefit of working with augmentations is that we only need to train a RISE model once with the augmented negatives.
The augmentations can be done in an offline manner, and thus reused for multiple experiments.
Once we have generated these augmentations, we then need to train a RISE model once on a given dataset augmented with negatives, and the resulting model will be our final model to be used for evaluating summaries.

\subsubsection{Model Negatives}

Model negatives are example negatives mined from a dataset.
The goal is to find negative examples that are similar to a reference summary, so that the model can learn to better differentiate these similar summaries.

To create this set of model negatives, we first finetune a RISE model on a dataset, and then we can mine within a dataset for similar summaries for each source document.
To do so, we encode all the documents and summaries, then for each document, find the top $n$ most similar summaries (excluding the associated summary of a document).
After we have finished mining for negatives, we then need to train a second model with the dataset that now contains the model negatives.
Once we have trained this second model, we can then use it for evaluating summaries.

The benefits of model negatives is that one is not dependent on needing to create methods for augmenting data.
It can also find similar summaries for a given reference that would not be achievable via augmentation, thus providing a model with a broader source of negatives.
The drawbacks though of model negatives is that one is required to train a model twice, once to be used for negative mining, and a second time for the final model to be used for evaluation.

\subsubsection{Combining Lexical and Model Negatives}

It is possible to combine the two above approaches for obtaining negatives.
Doing so would allow us to leverage the strengths of both types of negatives.

To do so, we first finetune a RISE model on the lexical negatives.
This resulting model is then used for the above mining process to find model negatives.
These model negatives are combined with the lexical negatives, creating a larger negative set for each example.
Then we finetune a final model on the combined dataset.

\section{Results}

We use SummEval \cite{fabbri-2021} to evaluate how well our approach correlates with human evaluations.
SummEval is a collection of human annotations on the quality of 16 models and their outputs for 100 examples from the CNN / Daily Mail task \cite{nallapati-etal-2016-abstractive}. 
As with past approaches, we focus on the annotations made by expert annotators, and use Kendall’s tau for system-level correlation.

As described in their work, there are four criteria used for human judgements:

\begin{itemize}
    \item Coherence -- the collective quality of the sentences in the summary.
    \item Consistency -- the consistency of the facts between the source and summary.
    \item Fluency -- the quality of the individual sentences in the summary.
    \item Relevance -- the selection of important content in the summary from the source.
\end{itemize}

\subsection{Methodology}

For our various experiments, we make use of T5 (specically version T5.1.1) when working with shorter contexts, LongT5 for longer contexts, and mT5 for multilingual tasks.
For T5 and mT5, we use input lengths of 4,096 for the input document, and for LongT5 we use input lengths of 16,384.
For all variants, we use a input length of 512 for the summary.

All models were trained for 30,000 steps; a batch size of 64; and the same T5 default learning rate with warmup steps set to 1500, base learning rate set to 0.001, and a decay factor of \num{7e-5}.
All models were also trained on the full training set for each respective dataset, with exception to Section \ref{sec:dataset_sizes}, in which we trained on partial datasets.

For exploration, we make use of a variety of abstractive summarization datasets.
This allows us to see how well RISE works whether the task is included or not within the training.
The training sets included are 
CNN / Daily Mail \cite{nallapati-etal-2016-abstractive},
Multi-News \cite{fabbri-etal-2019-multi},
arXiv \cite{cohan-etal-2018-discourse},
PubMed \cite{cohan-etal-2018-discourse},
BigPatent \cite{sharma-etal-2019-bigpatent},
SAMSum \cite{gliwa-etal-2019-samsum},
Reddit TIFU \cite{kim-etal-2019-abstractive},
and MLSUM \cite{scialom-etal-2020-mlsum}.

These datasets also allow us to explore situations of both short and long contexts.
CNN / Daily Mail, SAMSum, Reddit TIFU, and Multi-News can be used for training models of shorter context, while arXiv, PubMed, BigPatent and Multi-News\footnote{As reported by \citet{guo-etal-2022-longt5}, Multi-News when tokenized has on average 1,902 tokens and 4,853 at $90^{th}$ percentile, thus can be used with T5 when input limit is set to 4096 and also for LongT5 with its longer limits of 16k.} can be used for longer context.
MLSUM is used for testing on multilingual tasks.
Unless otherwise noted, all results are shown for Large-sized models.

For generating augmentations, specifically noun entities, we make use of spaCy v3.0\footnote{\url{https://spacy.io/}}, and for swapping antonyms, we make use of NLTK v3.7\cite{loper-bird-2002-nltk}.
For swapping noun entities, each entity noun seen in a summary example will be swapped with 50\% chance.
For dropping words or dropping characters, we drop at 20\% chance.
These random values were chosen arbitrarily and not further optimized.

All results shown are for the four human metrics of coherence, consistency, fluency, and relevance, along with the average of these metrics, making it easier to compare approaches.

\subsection{SummEval Comparisons}

Table \ref{tab:eval_results} show the results of our work with past approaches:
ROUGE \cite{lin-2004-rouge},
\textsc{chrF} \cite{popovic-2015-chrf},
SMS \cite{clark-etal-2019-sentence},
BARTScore \cite{yuan-2021-bartscore},
SMART \cite{amplayo-2022-smart},
BLEURT \cite{sellam-etal-2020-bleurt},
BERTScore \cite{zhang-2020-bertscore},
$Q^{2}$ \cite{honovich-etal-2021-q2},
T5-ANLI \cite{honovich-etal-2022-true-evaluating},
and PRISM \cite{thompson-post-2020-automatic}.
The scores shown for these past models are those reported by the recently-published SMART paper, in which we use the same methodology for evaluation.
We have grouped the approaches depending on whether they are reference-dependent or reference-free metrics, i.e., if they need a reference in order to be able to evaluate a score for a generated summary.

We show results from three of our RISE variants -- those trained on CNN/DM , Multi-News, and SamSUM.
We show those trained with a mixture of lexical negatives composed of 5 negatives with swapped entities and 5 negatives with randomly dropped words.
This configuration was shown to show strong performance and able to transfer well across datasets, as explained later in Section \ref{sec:ablations}.
These models are all Large-size.

As can be seen, all three variants show high correlation scores, particularly for consistency, fluency and relevance.
As expected, finetuning on in-domain data, in this case on CNN/DM, showed the strongest results.
Notably, comparing with other models that fine-tuned on CNN/DM, e.g., BARTScore-CNN, RISE achieves an absolute improvement of +16.7 points on the average metrics.

Additionally, we can see that RISE also performs well when finetuned on other, out-of-domain data, showing how well the model transfers to new summarization datasets.
While Multi-news is a bit similar to CNN/DM, in that both are in the domain of news articles, SamSUM is a dialogue summarization corpus and RISE stills transfer well when tested again the CNN/DM-based SummEval dataset.

Comparing with other metrics, first examining the similar reference-free metrics, we can see RISE performs more strongly than any past approach.
This is important when working in domains where one may want to evaluate new inputs that do not have a reference summary.
RISE also compares well with reference-dependent metrics, performing slightly better than the best metric SMART.

\begin{table}
\small
\centering
\begin{tabular}{lccccc}
\toprule
Metric & Coh & Con & Flu & Rel & Avg \\
\midrule
\multicolumn{6}{c}{\textit{Reference-dependent metrics}} \\
\midrule
ROUGE-1 & .350 & .550 & .527 & .583 & .503 \\
ROUGE-2 & .233 & .600 & .494 & .433 & .440 \\
ROUGE-L & .117 & .117 & .259 & .350 & .211 \\
BLEU & .217 & .050 & .326 & .383 & .244 \\
\textsc{chrF} & .350 & .617 & .561 & .550 & .519 \\ 
BERTScore & .333 & -.030 & .142 & .200 & .161 \\
MoverScore & .217 & -.050 & .259 & .350 & .194 \\
BLEURT & .533 & .200 & .410 & .467 & .403 \\
SMS & .267 & .600 & .360 & .400 & .407 \\
SMART-1 & .433 & .667 & .644 & .667 & .603 \\
SMART-2 & .417 & .750 & .628 & .583 & .594 \\
SMART-L & .567 & .567 & .611 & .733 & \textbf{.619} \\
\midrule
\multicolumn{6}{c}{\textit{Reference-free metrics}} \\
\midrule
PRISM & .233 & .600 & .360 & .367 & .390 \\
T5-ANLI & .250 & .583 & .544 & .517 & .473 \\
BARTScore & .350 & .617 & .494 & .450 & .478 \\
BARTScore\scriptsize+CNN\small & .550 & .317 & .594 & .583 & .511 \\
$Q^{2}$ & .250 & .750 & .577 & .450 & .507 \\
RISE$_{Multi-News}$ & .533 & .733 & .711 & .700 & .669 \\
RISE$_{SamSUM}$ & .533 & .700 & .678 & .700 & .653 \\
RISE$_{CNN}$ & .533 & .733 & .745 & .700 & \textbf{.678} \\
\bottomrule
\end{tabular}
\caption{
Results comparing past approaches with some of the RISE variants.
For the RISE variants, these are Large-sized models finetuned on a given dataset using lexical negatives, composed of a mixture of 5 summaries with swapped entities and 5 summaries with randomly dropped words for each dataset example.
SMART metrics are those reported when using BLEURT as the string matcher.
}
\label{tab:eval_results}
\end{table}

\subsection{Ablations}\label{sec:ablations}

We explore four ablations to see their impact on RISE -- impact of lexical and model negatives, size of the pre-trained model, datasets used for finetuning, and the size of the datasets used.
All ablation experiments are done using the Large-size model, with exception to Section \ref{sec:model_sizes} in which we explore the impact of model sizes.

\subsubsection{Lexical and Model Negatives}

We first look at how lexical and model negatives impact the performance of RISE.

Table \ref{tab:eval_augs} shows results of looking at different augmentations, focusing on CNN/DM, SamSUM, and Multi-News for the task.
As can be seen, swapping entity nouns and randomly dropping words perform the best as stand-alone augmentations for both tasks.
Combining the two (i.e., having 5 of each type of augmentation) results in even stronger performance.

\begin{table}
\small
\centering
\begin{tabular}{lccccc}
\toprule
Augmentations & Coh & Con & Flu & Rel & Avg \\
\midrule
\multicolumn{6}{c}{\textit{Multi-News}} \\
\midrule
SE & .467 & .733 & .678 & .667 & .636 \\
SW & .350 & .617 & .561 & .550 & .519 \\
DW & .467 & .733 & .678 & .667 & .636 \\
DC & .367 & .633 & .577 & .567 & .536 \\
SA & .400 & .633 & .577 & .600 & .553 \\
SE+DW & .533 & .733 & .711 & .700 & .669 \\
\midrule
\multicolumn{6}{c}{\textit{SamSUM}} \\
\midrule
SE & .600 & .633 & .644 & .633 & .628 \\
SW & .333 & .633 & .544 & .567 & .519\\
DW & .417 & .750 & .661 & .650 & .619 \\
DC & .317 & .617 & .527 & .550 & .503 \\
SA & .383 & .683 & .594 & .617 & .569 \\
SE+DW & .533 & .700 & .678 & .700 & .653 \\
\midrule
\multicolumn{6}{c}{\textit{CNN/DM}} \\
\midrule
SE & .733 & .467 & .644 & .633 & .619 \\
SW & .400 & .667 & .577 & .567 & .553 \\
DW & .467 & .733 & .711 & .667 & .644 \\
DC & .417 & .650 & .561 & .550 & .544 \\
SA & .483 & .550 & .494 & .483 & .503 \\
SE+DW & .533 & .733 & .745 & .700 & .678 \\
\bottomrule
\end{tabular}
\caption{Results of different augmentations.
SE is for swapping entities with source, SW is for randomly swapping words, DW is for randomly dropping words, DC is for randomly dropping characters, and SA is for randomly swapping words with their antonyms.
}
\label{tab:eval_augs}
\end{table}

Table \ref{tab:eval_augs_mining} shows comparisons with the same datasets when looking at either no negatives (i.e., only relying on in-batch negatives), using lexical negatives, using model negatives, or combining lexical and model negatives.
As can be see, lexical negatives have the largest impact for all task domains.
More surprisingly is that with working with the combined negatives, CNN/DM performs better while Multi-News and SamSUM performs worse.
We believe this is due to the model negatives helps the model when focused on the same task, which in turn makes the model less transferable to other domains.
While model negatives by themselves do not show as strong of a performance as lexical negatives, they can still be valuable if it is too expensive to create lexical negatives, or if one wants to train a model that is focused on a given domain and does not need the model to transfer to other domains.

\begin{table}
\small
\centering
\begin{tabular}{lccccc}
\toprule
Negatives & Coh & Con & Flu & Rel & Avg \\
\midrule
\multicolumn{6}{c}{\textit{Multi-News}} \\
\midrule
None & .250 & .550 & .494 & .450 & .436 \\
SE+DW & .533 & .733 & .711 & .700 & .669 \\
Mined 5 & .367 & .700 & .577 & .533 & .544 \\
Mined 10 & .367 & .667 & .577 & .533 & .536 \\
Mined 15 & .417 & .683 & .594 & .583 & .569 \\
Mined 20 & .383 & .650 & .594 & .583 & .533 \\
SE+DW+Mined 5 & .517 & .717 & .695 & .650 & .644 \\
\midrule
\multicolumn{6}{c}{\textit{SamSUM}} \\
\midrule
None & .283 & .550 & .494 & .483 & .453 \\
SE+DW & .533 & .700 & .678 & .700 & .653 \\
Mined 5 & .333 & .600 & .544 & .533 & .503 \\
Mined 10 & .283  & .583 & .494 & .517 & .469 \\
Mined 15 & .300 & .600 & .510 & .533 & .486 \\
Mined 20 & .317 & .583 & .527 & .517 & .486 \\
SE+DW+Mined 5 & .483 & .750 & .661 & .683 & .644 \\
\midrule
\multicolumn{6}{c}{\textit{CNN/DM}} \\
\midrule
None & .417 & .683 & .561 & .583 & .561 \\
SE+DW & .533 & .733 & .745 & .700 & .678 \\
Mined 5 & .433 & .700 & .611 & .633 & .594 \\
Mined 10 & .450 & .683 & .628 & .650 & .603 \\
Mined 15 & .367 & .667 & .544 & .567 & .536 \\
Mined 20 & .450 & .717 & .628 & .617 & .603 \\
SE+DW+Mined 5 & .550 & .717 & .762 & .750 & .695 \\
\bottomrule
\end{tabular}
\caption{Results of using no negatives (None), using augmentation, mining for negatives, and combining augmentations with mined negatives.
}
\label{tab:eval_augs_mining}
\end{table}

\subsubsection{Model Sizes}\label{sec:model_sizes}

Table \ref{tab:eval_base_size} shows results when looking at Base and Large model sizes.
The advantage of using Base is that, being smaller, it requires less computation for evaluating summaries.
And as can be seen, Base models tend to perform adequate enough if focused on in-domain data, but do not transfer as well to other domains when compared to Large.
Even though Large models do come with overhead of requiring more computation, we believe it is worth the trade off to get the much stronger performance, while at the same time not being too prohibitive in computation costs.

\begin{table}
\small
\centering
\begin{tabular}{lccccc}
\toprule
Size & Coh & Con & Flu & Rel & Avg \\
\midrule
\multicolumn{6}{c}{\textit{Multi-News}} \\
\midrule
Base & .367 & .333 & .276 & .267 & .311 \\
Large & .533 & .733 & .711 & .700 & .669 \\
\midrule
\multicolumn{6}{c}{\textit{SamSUM}} \\
\midrule
Base & .483 & .417 & .393 & .383 & .419  \\
Large & .533 & .700 & .678 & .700 & .653 \\
\midrule
\multicolumn{6}{c}{\textit{CNN/DM}} \\
\midrule
Base & .600 & .433 & .477 & .467 & .494 \\
Large & .533 & .733 & .745 & .700 & .678 \\
\bottomrule
\end{tabular}
\caption{Results of comparing Base and Large model sizes.}
\label{tab:eval_base_size}
\end{table}

\subsubsection{Datasets}

We next look at the impact of training RISE on different datasets.

Table \ref{tab:eval_t5_datasets} shows results when running on different datasets that are applicable with a T5 model.
This also includes mixing datasets, to see whether we can benefit from mixtures.
As the results first show, RISE shows it can transfer well across different datasets.
While Multi-News is similar to the CNN/DM dataset used in SummEval, the other datasets SamSUM and the two Reddit variants are rather different.
Suprisingly though, the model does not perform as well when mixing datasets -- RISE works better if trained on just a single dataset, either for transfering to a new domain or within a domain.

\begin{table}
\small
\centering
\begin{tabular}{lccccc}
\toprule
Dataset & Coh & Con & Flu & Rel & Avg \\
\midrule
Multi-News & .533 & .733 & .711 & .700 & .669 \\
SamSUM & .533 & .700 & .678 & .700 & .653 \\
Reddit - TLDR & .500 & .733 & .711 & .667 & .653 \\
Reddit - Title & .467 & .767 & .695 & .633 & .640 \\
CNN/DM & .533 & .733 & .745 & .700 & .678 \\
Mixed - CNN  & .567 & .633 & .611 & .600 & .603 \\
Mixed + CNN  & .533 & .733 & .711 & .667 & .661 \\
\bottomrule
\end{tabular}
\caption{Results of various datasets when tested on SummEval.
Mixed datasets are mixing the datasets within the table, either without or with CNN/DM.
These are all reporting results when using lexical negatives of 5 with swapped entities and 5 with randomly-dropped words.}
\label{tab:eval_t5_datasets}
\end{table}

Table \ref{tab:eval_longt5_datasets} shows results when training on datasets with LongT5.
As shown, RISE with LongT5 does perform worse than that of T5 -- this can be expected as T5's full attention is better suited (this is also supported in the original LongT5 paper, where LongT5 on CNN/DM did not perform as well against other datasets when compared to training on long-context datasets).
Despite the weaker performance, the model still performs comparable to other reference-free metrics, many of which we do not know how well they would scale up to long-context datasets.
In terms of individual models, the model trained on Multi-News did slightly worse despite being news-related.
This may be that LongT5 is better able to capture the full input, thus handling multiple documents, which differs from the other datasets presented and CNN/DM, in that they are all of a single document input.

\begin{table}
\small
\centering
\begin{tabular}{lccccc}
\toprule
Dataset & Coh & Con & Flu & Rel & Avg \\
\midrule
Multi-News & .483 & .350 & .427 & .383 & .411 \\
arXiv & .400 & .633 & .544 & .533 & 528 \\
PubMed & .517 & .617 & .594 & .583 & .578 \\
BigPatent & .517 & .450 & .460 & .417 & .461\\
\bottomrule
\end{tabular}
\caption{Results of various datasets using LongT5 when tested on SummEval.
These are all reporting results when using negatives of 5 with swapped entities and 5 with randomly-dropped words.}
\label{tab:eval_longt5_datasets}
\end{table}

Table \ref{tab:eval_multi} shows the results of finetuning on the multilingual summarization task MLSUM.
We finetuned on 3 of the languages within this task, Spanish (ES), German (DE), and French (FR).
As can be seen, despite having been trained on other languages, RISE still shows strong correlation with human metrics when applied to an English dataset.

\begin{table}
\small
\centering
\begin{tabular}{lccccc}
\toprule
Dataset & Coh & Con & Flu & Inf & Avg \\
\midrule
MLSUM-ES & .483 & .683 & .661 & .650 & .619 \\
MLSUM-DE & .450 & .717 & .628 & .583 & .594 \\
MLSUM-FR & .550 & .683 & .695 & .650 & .644 \\
\bottomrule
\end{tabular}
\caption{Results of multilingual datasets when tested on SummEval. These are all reporting results when using negatives of 5 with swapped entities and 5 with randomly-dropped words.}
\label{tab:eval_multi}
\end{table}

\subsubsection{Dataset Sizes}\label{sec:dataset_sizes}

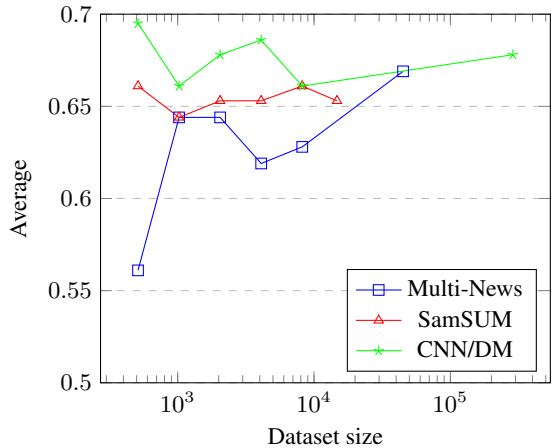
\begin{figure}
\begin{tikzpicture}
\centering
\small
\begin{axis}[
    title={},
    xlabel={Dataset size},
    ylabel={Average},
    ymin=.5, ymax=.7,
    legend pos=south east,
    ymajorgrids=true,
    grid style=dashed,
    xmode=log,
]

\addplot[
    color=blue,
    mark=square,
    ]
    coordinates {
    (512,.561)(1024,.644)(2048,.644)(4096,.619)(8192,.628)(44972,.669)
    };
    
\addplot[
    color=red,
    mark=triangle,
    ]
    coordinates {
    (512,.661)(1024,.644)(2048,.653)(4096,.653)(8192,.661)(14732,.653)
    };

 \addplot[
    color=green,
    mark=star,
    ]
    coordinates {
    (512,.695)(1024,.661)(2048,.678)(4096,.686)(8192,.661)(287113,.678)
    };
    \legend{Multi-News,SamSUM,CNN/DM}

\end{axis}
\end{tikzpicture}
\caption{Results of looking at reduced amount of data for training.
All three datasets were trained with reduced sizes of 512, 1024, 2048, 4096, and 8192, compared to the full set size of 44,972 for Multi-News, 14,732 for SamSUM, and 287,113 for CNN/DM. 
}
\label{fig:eval_size_count}
\end{figure}

As a final ablation, we examined how well the model performs even with a reduced amount of data.
Table \ref{fig:eval_size_count} shows the results of these experiments.
As can be seen, the model still does well with reduced-data for the 3 datasets we trained upon.
Only when trained on Multi-News with 512 examples do we see a bit of drop off in performance.
This indicates that the model can learn well in domains where one might not have much data as with the full datasets used in this paper.

\subsection{arXiv and GovReport Comparisons}

As an additional comparison, we look at longer documents annotated by \citet{long-summ-eval}.
In this work, the authors had human raters annotating various summaries of models on arXiv and GovReport \cite{huang-etal-2021-efficient}.
They were looking at two metrics, relevance and factual consistency.
For these evaluations, we use Spearmann rank correlation.

Table \ref{tab:eval_long_summ} shows the results of comparing RISE with these human evaluations.
We compared RISE with the results of BARTScore and SMART.
For the arXiv dataset, we can see that applying our various models trained on LongT5 show higher average correlations.

The GovReport dataset is a bit different than past datasets, in that its summaries are much longer.
When tokenized, the median summary for arXiv is 249, while GovReport is 657.
To allow for a model that can handle these longer summaries better, we finetuned RISE on GovReport.
As shown in Table \ref{tab:eval_long_summ} in the bottom half, RISE is then able to perform well in correlation with the human evaluations.
More importantly, using only lexical data puts it on par with SMART, but when we add the model negatives (creating a combined lexical and model negatives data), this results in much stronger correlations.

\begin{table}
\small
\centering
\begin{tabular}{lccc}
\toprule
Size & Rel & Fac & Avg \\
\midrule
\multicolumn{4}{c}{\textit{arXiv}} \\
\midrule
BARTScore & .12 & .24 & .17 \\
SMART & .45 & .38 & .41 \\
\midrule
RISE$_{arXiv}$ & .75 & .33 & .54 \\
RISE$_{PubMed}$ & .74 & .66 & .70 \\
RISE$_{BigPatent}$ & .67 & .50 & .59 \\
RISE$_{Multi-News}$ & .57 & .71 & .64 \\
\midrule
\multicolumn{4}{c}{\textit{GovReport}} \\
\midrule
BARTScore & .25 & .06 & .12 \\
SMART & .31 & .26 & .28 \\
\midrule
RISE$_{GovReport}$ - lexical negs & .39 & .15 & .27 \\
RISE$_{GovReport}$ - combined negs & .61 & .22 & .42 \\
\bottomrule
\end{tabular}
\caption{Results comparing past approaches with RISE on longer documents of arXiv and GovReport.
Note that the metrics here are using Spearmann rank correlation.}
\label{tab:eval_long_summ}
\end{table}

\subsection{Overall Recommendations}

There are many ways to train RISE, and different model architectures one can use.

Given the results, we first recommend using architecture that matches the length and types of inputs.
For short inputs, it is best to use a model trained on T5; for multilingual inputs, it is best to use a model trained on mT5; and for long inputs, it is best to use a model trained on LongT5.

What type of data to use also depends how one is expecting to use the model for evaluation.
If one needs a model only focused on a given domain, then training with both lexical and model negatives gives best results.
If one needs a model that can transfer to other domains, then it is best to use just lexical negatives.

We have released checkpoints from many of the models presented in this paper, allowing for one to reuse this work for their own evaluations on datasets commonly used in summarization.

\section{Conclusion}

We have presented our new model RISE for evaluating text summaries.
As the results show, RISE has strong correlation with human evaluations.
Being a reference-free metric, it can be used in new domains where generating golden summaries may be prohibitive.
And while RISE shows strong correlation with human evaluations, we do not view RISE as a replacement of other metrics.
Instead, we view it as complementary, especially to reference-dependent metrics such as ROUGE, \textsc{chrF}, and SMART.

Summarization evaluation continues to be a challenging problem.
Leveraging data from within the domain can help though with calibration of evaluation metrics.
BARTScore had earlier touched upon this, and RISE further helps show the importance of this.
We hope this will help spur future research in how we can use domain data to improve such metrics.

One of the benefits of RISE not explored but left for future work is how RISE can be tailored to address specific needs.
Since RISE depends on contrastive learning, one can create negatives that reflect characteristics they want to specifically evaluate. Another area of future work is looking at the possibility of using RISE in other generative domains outside of summarization, such as dialogue systems.
Many of the techniques here for training can easily be applied to other domains, including creation of lexical and model negatives.

\section*{Limitations}

As with many other model-based metrics, RISE is best suited for evaluating offline due to the expensive nature of inferring with a large model.
It is not as well suited as other metrics like ROUGE or BLEU for evaluating during training or fine-tuning. We leave the exploration of using RISE for evaluation-in-the-loop kind of training for summarization models future work.

Additionally, as with other model-based metrics, it is possible that the models may have seen some of the data during pretraining as is in the eval datasets.
We do not think it would be too significant, as the pretraining task (for T5/mT5 for example) is rather different than a summarization task and, more importantly, it does not include the gold reference.
Thus the model would not be able to make such a connection easily despite having seen the data.

We chose to work with the T5-family of models due to the ease-of-use for others to implement and improve upon our ideas.
We would expect our ideas to work just as well with other models, such as BART, mBART, Longformer, etc.

Following recent works, we have studied the evaluation based on the SummEval benchmark \cite{fabbri-2021}. In the future, we may want to build other benchmarks that covers more domains and languages to compare different methods.

\section*{Ethics Statement}

RISE is built upon pre-trained language models. Any biases within these models may possibly influence the scoring of summarization models, in that it is possible biases may cause the models to rate one summary better than another.

\bibliography{acl_latex}
\bibliographystyle{acl_natbib}

\end{document}